\documentclass{article}

\usepackage[nonatbib, final]{nips_2017}

\usepackage[utf8]{inputenc} % allow utf-8 input
\usepackage[T1]{fontenc}    % use 8-bit T1 fonts
\usepackage{hyperref}       % hyperlinks
\usepackage{url}            % simple URL typesetting
\usepackage{booktabs}       % professional-quality tables
\usepackage{amsfonts}       % blackboard math symbols
\usepackage{amsmath}
\usepackage{nicefrac}       % compact symbols for 1/2, etc.
\usepackage{microtype}      % microtypography
\usepackage{eurosym}
\usepackage{graphicx}

\title{Subgroup Identification and Interpretation with Bayesian Nonparametric Models in Health Care Claims Data}

\author{
  Christoph F. Kurz \\
  Helmholtz Zentrum M\"unchen, Germany\\
  Institute of Health Economics and Health Care Management\\
  \texttt{christoph.kurz@helmholtz-muenchen.de} \\
  \And 
  Laura A. Hatfield \\
  Harvard Medical School \\
  Department of Health Care Policy \\
}

\begin{document}
% \nipsfinalcopy is no longer used

\maketitle

\begin{abstract}
Inpatient care is a large share of total health care spending, making analysis of inpatient utilization patterns an important part of understanding what drives health care spending growth. Common features of inpatient utilization measures include zero inflation, over-dispersion, and skewness, all of which complicate statistical modeling. Mixture modeling is a popular approach that can accommodate these features of health care utilization data. In this work, we add a nonparametric clustering component to such models. Our fully Bayesian model framework allows for an unknown number of mixing components, so that the data determine the number of mixture components. %In simulation studies, we show that this model finds the true number of mixture components more accurately than using information criteria for model selection. 
When we apply the modeling framework to data on hospital lengths of stay for patients with lung cancer, we find distinct subgroups of patients with differences in means and variances of hospital days, health and treatment covariates, and relationships between covariates and length of stay.
\end{abstract}

\section{Introduction}
Inpatient hospital services account for a small share of health care utilization, but the majority of total health care spending.~\cite{kashihara2009statistical} To understand the variation in this major component of health care expenditures, researchers have sought to identify patient subgroups with different utilization and spending patterns.~\cite{hingorani2013prognosis}

Health care resource use data are often non-negative, right-skewed, heavy-tailed, and multi-modal with a point mass at zero. Desirable analytical approaches for these data should be sufficiently powerful and flexible to accommodate all these features. Common generalized linear models (GLMs) for count data use the Poisson, geometric, and Negative Binomial distributions, which do not account for over-dispersion, zero inflation, and multi-modality.~\cite{greene1994accounting, hilbe2011modeling} More sophisticated models for counts include the zero-inflated or hurdle model~\cite{duan1983comparison} and finite mixture models (FMMs), also known as latent class models. FMMs identify groups of observations with similar outcomes using unsupervised clustering. FMM models may be implemented with count distributions, such as Poisson~\cite{pohlmeier1995econometric} and Negative Binomial~\cite{deb1997demand}, and may be augmented with a hurdle component for excess zeros.~\cite{bago2006latent} FMMs provide better fit than standard GLMs and the hurdle model.~\cite{deb1997demand} In addition, mixture models accommodate multi-modality and can even link mixture component prevalences to covariates.~\cite{mihaylova2011review}

Mixture models avoid the hurdle model's sharp dichotomy between users and non-users. 
A key question in mixture models is the optimal number of components. (Note that we use \emph{component}, rather than \emph{cluster}, to describe the subpopulations identified by FMMs.) Too many components may overfit the data and impair model interpretation, while too few components limit the flexibility of the mixture to approximate the true underlying data structure.
The number of components can be decided  \emph{ex ante}, by choosing a convenient and interpretable number such as two or three, or \emph{ex post}, by calculating models with different numbers of components and comparing their fit statistics.

Our model uses a Dirichlet prior (DP) for the mixing component.
In this fully Bayesian hierarchical mixture model, the optimal number of components is determined simultaneously with the model fit, and the model can incorporate prior information about the number of components. This one-stage process yields the ideal number of components and allows interpretation of each component. The potentially unbounded infinite mixture model avoids both over- and under-fitting by allowing the data to determine the optimal number of components.~\cite{rasmussen2000infinite} 

We are motivated by the desire to understand variation in days spent in hospital among patients diagnosed with lung cancer. Lung cancer is the most common cancer worldwide and a major cause of cancer-related mortality. Previous work has shown substantial heterogeneity in the patterns of health care utilization among patients with lung cancer.~\cite{schuler2017trajectories} To understand patient subpopulations defined by inpatient hospital days and effects of covariates on hospital days, we develop and apply Negative Binomial regression models that use a DP prior for nonparametric mixing of the components. Studies found that nonparametric models can outperform current methods, especially in the case of health claims data.~\cite{fellingham2015bayesian}

\section{Dirichlet Prior Mixture Regression for Count Responses (DP-NB)}
The basis of our model is a Negative Binomial regression model.
We extend this model to a mixture of Negative Binomial components indexed by $k=1,\ldots,K$. Each component is  parameterized by a mean $\boldsymbol{\mu}=\mu_1,...,\mu_K$, and a precision parameter $\boldsymbol{\psi}=\psi_1,...,\psi_K$. As before, we specify a regression model for the mean parameter $\mu_k =\text{exp}(\mathbf{x}\beta_{k})$. We define $\mathbf{z}=z_1,...,z_K$ as the component assignment variable, where each $z_k$ is a $K$-indicator vector. The component assignment depends on the mixing variable $\mathbf{c}=c_1,...,c_K$, with $\sum_{k=1}^K c_k=1$. Thus, the likelihood is:
$$
p(y\mid \mu,\psi) = \prod_{n=1}^N\prod_{k=1}^K \left( \text{NegBin}(y_n \mid \mu_k,\psi_k) \right)^{z_{nk}}
$$
We define the marginal over component assignments as a categorical distribution:
$$
p(\mathbf{z}\mid \mathbf{c})= \prod_{n=1}^N \prod_{k=1}^K c_{k}^{z_{nk}},
$$
and the prior over the $K$-simplex mixing vector as a Dirichlet distribution with hyperparameter $\alpha_0$:
$$
p(\mathbf{c}) = \text{Dir}(\mathbf{c}\mid\alpha_0) = \frac{\Gamma(K\alpha_0)}{\Gamma(\alpha_0)^K}\prod_{k=1}^{K}c_k^{\alpha_0-1},
$$
where $\Gamma(\cdot)$ is the Gamma function.
Using a Dirichlet prior for the distribution of component means in mixture models does not require one to specify the number of components. Instead, a concentration parameter $\alpha_0$ controls it implicitly. The Dirichlet prior serves as a nonparametric prior on the mixture components.
Finally, we specify priors for the regression coefficients and the precision parameters:
$$
\begin{aligned}
\beta_{kd}  &\sim \mathcal{N}(m_0,s_0),\\
\psi_k &\sim \mathcal{LN}(a_0,b_0).
\end{aligned}
$$
We chose the following weakly informative~\cite{gelman2006prior} hyperparameters:
$$
\alpha_0= 0.1,\quad m_0=0.0,\quad s_0=10.0,\quad a_0=0.0,\quad b_0=2.0.
$$
In addition to the usual regression parameters, this nonparametric mixture models produces several additional parameters of interest. For each mixture component $k$, we want to estimate the relative prevalence of the mixture component in the data and parameters of the mixture component's distribution, such as the mean, variance, and regression coefficients. The mixture weights $c_k$ are the probabilities associated with each component. In addition, for each observation, we want to estimate the mixture component from which it was most likely drawn, also called the component assignment $z$.

\section{AOK Data Set}
We analyzed health care billing claims provided by the AOK Research Institute. AOK covers around 30\% of the German resident population. The data set contains patient-level information on inpatient and outpatient diagnoses and procedures from 2009 to 2012, as well as service utilization. We used a study population previously derived from this data set consisting of patients with incident lung cancer in 2009. More information on the data set and the selection criteria can be found in~\cite{schwarzkopf2015cost}.

The outcome of interest was the total number of inpatient hospital days for each patient in the year after diagnosis. Inpatient hospital days are defined as the number of days from formal admission to hospital until discharge (i.e., hospital outpatient procedures do not count as hospital days), summed over all hospitalizations in a year. Admission and discharge on the same day count as one hospital day, so only individuals who were never admitted have zero hospital days. We included only individuals who survived for the full year, resulting in $N=7118$ individual observations. The mean number of hospital days is 44, with a maximum of 296 and 59 zero observations.

We included the following covariates in the model: age, sex, treatment type during the course of the disease (chemotherapy, radiation therapy, or surgery), number of other tumor sites at diagnosis, number of metastases at diagnosis, Charlson comorbidity index, and district type of residence (major city, urban district, rural district, or thinly populated rural district). The Charlson comorbidity index was calculated using ICD-10 codes as in~\cite{sundararajan2004new} with the slight modification of excluding the diagnosis of lung cancer out of the group ``solid tumor without metastation".

\section{Results\label{s:results}}
%\subsection{Component Identification in Simulated Data}
%
%
%\subsection{Mixture Regression Analysis of AOK Data}
For the AOK data set, the DP-NB finds 3 components as having the highest expected posterior mixture weights. In the following, we only show results based on this final model with 3 components. 

Component 1 contains only 6\% (420/7118) of all observations and corresponds to individuals who spend, on average, fewer days in hospital but with very high variance. The mode of hospital days for component 1 is 5 for the DP-NB and 7 for the DP-ZINB (zero-mode omitted). Component 2 is the largest, with 58\% (4091/7118) of individuals; they stay longer in hospital (the mode is at 24 days) with less variance. Component 3 comprises 37\% (2607/7118) of the population, and these patients have the most hospital days (mode at 39 days) and again high variance.

\begin{figure}[h!]
	\centering
	\includegraphics[width=0.99\textwidth]{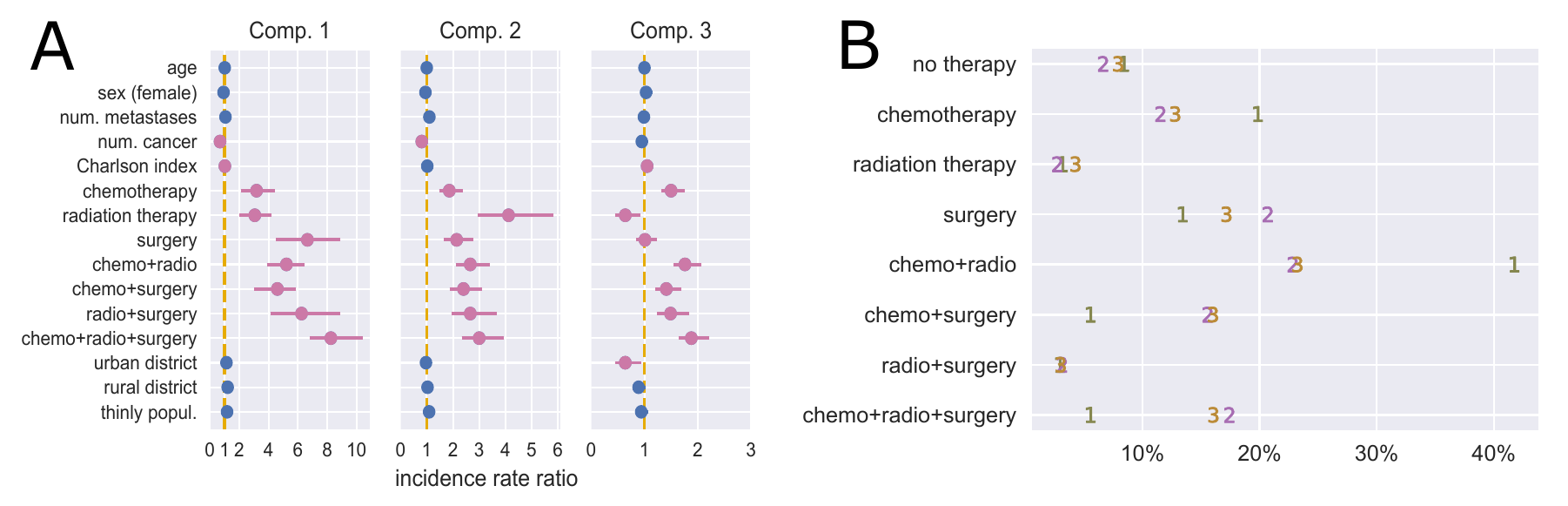}
	\caption{(A) DP-NB estimation results for all three components on the AOK data set. Parameter estimates are presented as incidence rate ratios and 95\% high probability density intervals. Intervals that exclude the 1 are highlighted in purple. Intercept is not shown.}\label{fig:combined}
\end{figure}

Figure~\ref{fig:combined} (A) shows $\beta_k$ parameter estimates for each component of the DP-NB as incidence rate ratios (IRRs) alongside the Bayesian high probability density intervals (HPDIs). These coefficients can be interpreted as the multiplicative increase in the expected number of hospital days for every one unit increase in the predictor. For example, in components 1 and 2, treatment is associated with more hospital days, across all modalities. The IRR for the combination of all three treatments is the largest, 8.3. That is, compared to patients who receive no treatment, those who receive chemotherapy, radiation, and surgery have 8.3 times as many expected hospital days. In general, all treatment combinations have higher IRRs in component 1 than in components 2 and 3. The only exception is radiotherapy, which has an IRR of 4.1 in component 2, higher than 0.6 in component 3, and 3.1 in component 1. The IRR for the number of metastases is decreasing over the components: 1.24, 1.10 in 2, and 0.99 in 3. On the other hand, the IRR for the number of multiple tumors is increasing from 0.69 to 0.81 to 0.95 across components 1, 2, and 3. In component 3, radiation is associated with fewer hospital days and surgery is null, whereas chemotherapy and combinations are associated with more hospital days. Demographic factors and baseline health were less strongly associated with hospital days. Age appears to have no relationship to hospital days (the IRRs are around 1.0 in all components), and sex has a very small estimate in component 1 (IRR of 0.93).
Regional factors are only important for individuals in component 3 and only for urban districts, where the IRR is 0.64.

When we use hard assignments to classify individuals into components according to the highest posterior probability, we see that treatment patterns are very different across components (see Figure~\ref{fig:combined} (B)). Chemotherapy plus radiation is the most common treatment in all components, but individuals in component 1 are far more likely to receive this combination (42\% compared to 23\% in components 2 and 3). Surgery alone is the second most common treatment in components 2 (22\%) and 3 (17\%), and chemotherapy alone is the second most common treatment in component 1 (20\%), but it is infrequent in 2 (12\%) and 3 (13\%). Compared to people in component 1, those in components 2 and 3 are more likely to receive chemotherapy combined with surgery or surgery and radiation. %Figure~\ref{fig:boxplot} shows greater comorbidity burden and slightly older age in component 3. The Charlson comorbidity burden increased across the components: median 1 (IRQ 0--3) in component 1; median 2  (IQR 0--3) in component 2; and median 2 (IQR 1--4) in component 3. Metastases were more common in patients in component 1 (median 1, IQR 0--1) than in components 2 and 3 (median 0, IQR 0--1).

\section{Discussion\label{s:discussion}}
This paper explores a Bayesian regression model for count data that combines nonparametric clustering with the advantages of finite mixture models. It avoids model selection and decision bias because all parameters, including the ideal number of mixture components, are estimated from the data. 

For the AOK data set, the DP models find three components of individuals with strikingly different distributions of hospital days and treatment patterns. 
In the treatment of lung cancer, surgery offers the best prospect of cure. If diagnosed at an early stage, it is possible to remove the tumor as a whole, such that  no further treatment is necessary. If the tumor is already bigger, surgical resection with chemotherapy and radiation is the treatment of choice. In metastatic lung cancer, palliative chemotherapy, possibly accompanied by radiation therapy for individual metastases, may alleviate symptoms and prolong survival.~\cite{gridelli2001quality}

Component 1 has the fewest hospital days on average. In this component, we find many patients with chemotherapy only, and chemotherapy in combination radiation therapy. This, and the lack of surgery, likely indicate that these patients were already in an advanced (metastatic) stage at diagnosis. For these patients, it is likely that therapy had a palliative intent with a focus on improving quality of life. In contrast, patients in components 2 and 3 were more likely to have surgery only, surgery and chemotherapy, and the combination of all three treatments. This indicates diagnosis at an earlier stage and more aggressive treatment. 
The treatments received by people in components 2 and 3 are quite similar, with only the proportion of surgery being slightly higher in 2 than in 3. However, the coefficients governing the relationships between treatment type and hospital days in these components are quite different. While radiation therapy is associated with significantly more hospital days in component 2, it has the opposite association in component 3. Moreover, the strong and positive association of surgery with hospital days in component 2 fades in component 3, where surgery has no relationship to hospital days. Together, these results suggest that people in components 2 and 3 get similar treatment combinations, but for different reasons.

There are several limitations to this study. First, DP mixture models present computational challenges. For example, care must be taken when fitting Bayesian mixture models to avoid the so-called "label-switching problem" caused by the model being invariant under permutations of the indices of the components (i.e., the indices of the model components may be permuted across chains). It is crucial to run multiple Markov chains, inspect the resulting posterior samples, and apply posterior checks, as we have done here.  In addition, because nonparametric mixture models allow inference over such a large parameter space (i.e., DP mixture models use a prior over essentially all distributions), posterior computation may be intractable. Neal~\cite{neal2000markov} provides an overview of the state-of-the-art sampling algorithms, but much progress has been made since then. In our applications, fitting this model to the AOK data took approximately 4 hours on a current quadcore CPU with 32GB RAM, compared to only 3 minutes to compute a finite mixture model with a pre-specified number of components. Further research should investigate how variational Bayesian methods could improve speed and how this affects the accuracy of the estimates.

In addition, interpretation is necessarily more difficult in complex models such as these. In the case of the DP-NB model, with three mixture components, the number of regression coefficients is three times the number of covariates. However, inference on multiple parameters simultaneously is relatively straightforward in Bayesian models, which is another advantage of this approach.

\small
\bibliographystyle{plain}
\bibliography{sample}

\end{document}